\documentclass[letterpaper, 12pt]{article}
\usepackage[margin=1in]{geometry}

\usepackage{generic}

\usepackage{amsmath}
\allowdisplaybreaks
\usepackage{xcolor}
\usepackage{footnote}
\usepackage{algorithm}
\usepackage{algpseudocode}
\usepackage{algorithmicx}
\usepackage{multirow} 
\usepackage{booktabs}
\usepackage{graphicx}
\usepackage{amssymb}
\usepackage{amsbsy}

\usepackage{amsthm}

\usepackage{array}
\usepackage{longtable}
\usepackage{epstopdf}
\usepackage{pbox}
\usepackage{breqn}
\usepackage{mathrsfs}
\usepackage{multicol}
\usepackage{supertabular}
\usepackage{enumerate}
\usepackage[colorinlistoftodos]{todonotes}
\usepackage{url}
\usepackage[justification=centering]{caption}
\usepackage{tabu}
\usepackage{subcaption}
\usepackage{graphics} 
\usepackage{epsfig} 

\usepackage{mathtools}

\usepackage{tikz}
\usepackage{tikz-network}

\usepackage{cases}
\usepackage{bm}
\usepackage{cite}

\usepackage{aligned-overset}

\newtheorem*{thm*}{Theorem}

\theoremstyle{definition}

\newcommand{\thistheoremname}{}
\newtheorem*{genericthm*}{\thistheoremname}
\newenvironment{namedthm*}[1]
  {\renewcommand{\thistheoremname}{#1}%
  \begin{genericthm*}}
  {\end{genericthm*}}

\title{\LARGE
Physics-informed Machine Learning for Calibrating Macroscopic Traffic Flow Models
}

\author{Yu Tang, Li Jin and Kaan Ozbay
\thanks{This work was in part supported by US NSF Award CMMI-1949710, USDOT Award \# 69A3551747124 via the C2SMART Center, NYU Tandon School of Engineering, SJTU UM Joint Institute, and J. Wu \& J. Sun Endowment Fund.}
\thanks{Y. Tang is with C2SMART Center, Department of Civil \& Urban engineering, Tandon School of Engineering, New York University, USA. L. Jin is with the UM Joint Institute and with the Department of Automation, Shanghai Jiao Tong University, China. K. Ozbay is with C2SMART Center, Department of Civil \& Urban engineering, Tandon School of Engineering, New York University, 11201, USA.
(emails: tangyu@nyu.edu, li.jin@sjtu.edu.cn, kaan.ozbay@nyu.edu).}
}

\date{Extended Abstract}

\begin{document}
\maketitle


{\bf Key Words}:
Physics-informed machine learning, traffic flow models, parameter identification.

\section{Introduction}

\subsection{Motivation}
Macroscopic traffic flow models have been shown to  be capable of reproducing congestion propagation and explaining complicated phenomena, such as capacity drops \cite{khoshyaran2015capacity} and stop-and-go waves \cite{laval2010mechanism}. They provide a solid foundation for the performance analysis of traffic systems \cite{huang2020scalable, shi2021constructing} and control design for freeway management \cite{gomes2006optimal, papamichail2010coordinated}. However, before macroscopic models are put into practice, they should be carefully calibrated to accurately replicate real-life complications briefly mentioned above.

\begin{figure}[htbp]
    \centering
    \includegraphics[width=0.7\linewidth]{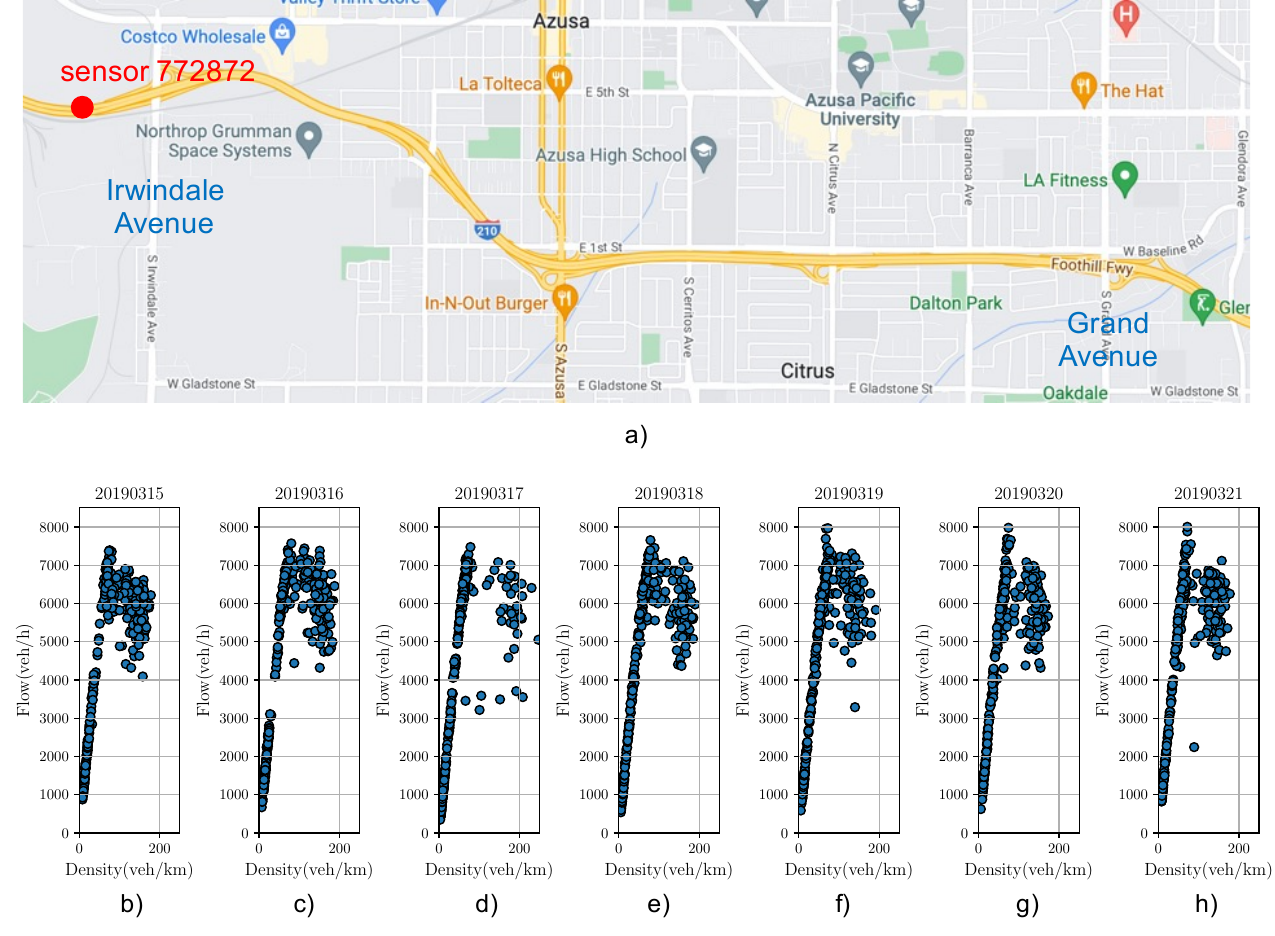}
    \caption{Fundamental diagrams varying day by day: a) a stretch of Interstate 210 Eastbound (I-210 E) from Irwindale Ave. to Grand Ave. in California, b)-h) seven days fundamental diagrams revealed by sensor 772872. Clearly, traffic flow parameters, such as capacity, are subject to day-to-day changes. Obvious capacity drops happened on 2019-03-19, 2019-03-20 and 2019-03-21.}
    \label{fig_motivating1}
\end{figure}

Extensive studies have been devoted to the calibration of traffic flow models \cite{spiliopoulou2014macroscopic,spiliopoulou2017macroscopic,mohammadian2021performance,wang2022macroscopic}. They mainly utilized optimization algorithms to determine model parameters over a certain period (i.e., morning or evening peaks) with multiple  days' data, but only a few of them, with the goal of testing for parameter transferability \cite{wang2022macroscopic}, validated the calibrated models for the same period but on different days. This transferability can be poor since road traffic is prone to perturbations induced by demand variations, weather conditions, driving behavior, and so on, as illustrated by the example in Figure~\ref{fig_motivating1}: our data analysis indicates that daily re-calibration is necessary for quantifying parameter uncertainties.

\begin{figure}[htbp]
    \centering
    \includegraphics[width=0.5\linewidth]{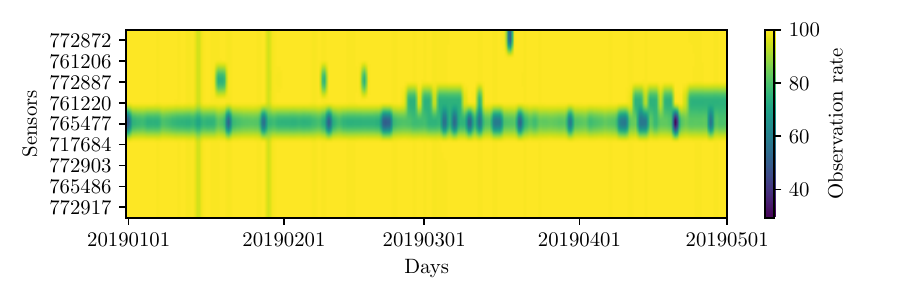}
    \caption{Observation rates of sensors along I-210 E from Irwindale Ave. to Grand Ave. in California, which stand for percentages of observed data, e.g., zero percent indicates sensor breakdown without any measurements.}
    \label{fig_motivating2}
\end{figure}

Admittedly, one can apply optimization-based approaches to day-to-day calibration, but this can be cumbersome in practice. First, these methods bring about heavy computation costs when being applied to long-term modeling up to several months or years. They typically involve non-convex and even non-smooth problems, and it is hard to solve them using optimization, let alone to repeat the calibration procedures for each day. The second problem arises from data quality. Traditional traffic sensors, such as widely-used inductive loops, are infamously unreliable. For instance, it is reported that only around 64\% of detectors work any given day in the California's freeway system \cite{PeMS}. Figure~\ref{fig_motivating2} illustrates practical observation rates of induction loops that are closely related with how many convincing observations are collected. Clearly, day-to-day optimization methods are vulnerable to the fluctuation in data quality. 
However, limited research discussed how to fulfill parameter identification on corrupted data in a robust way.

In response to the above challenges, we develop a novel physics-informed, learning-based approach of identifying traffic flow parameters across days, including capacity, free flow speed, jam density and congestion wave speed. The proposed method belongs to the unsupervised machine learning category since the actual values of parameters are unknown in advance. We inform our machine learning model of physics knowledge about traffic flows so that it can achieve the calibration even without ground truths of the parameters. Once well trained, our method is expected to yield reasonable parameter values given boundary conditions and traffic measurements with possibly missing values. It provides efficient calibration over massive periods with robustness to incomplete data. 

\begin{figure}[htbp]
    \centering
    \includegraphics[width=\linewidth]{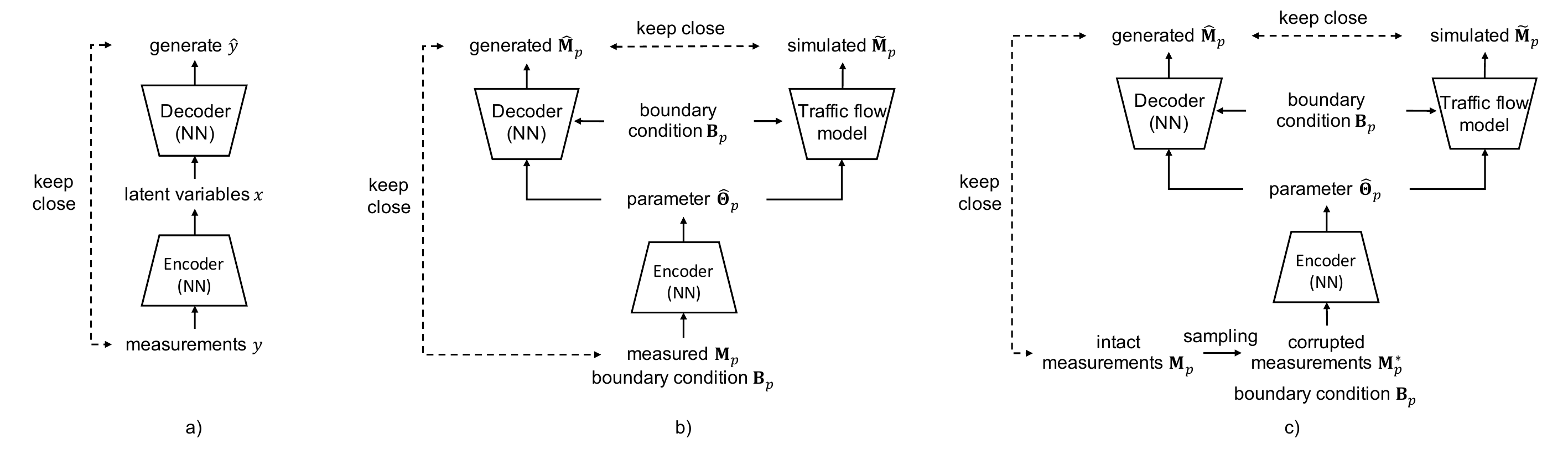}
    \caption{Frameworks of autoencoder and our extensions for physics-informed, learning-based calibration: a) deep autoencoder comprising two neural networks (NNs), one acting an encoder and the other one acting as a decoder, b) autoencoder-based parameter identification, c) denoising autoencoder-based parameter identification.}
    \label{fig_extension}
\end{figure}


\subsection{Our contributions}
We try to address two main questions:
\begin{enumerate}
    \item[(i)] How can we train an encoder, given perfect freeway measurements, to yield appropriate traffic parameters that are comparable to those calibrated by traditional methods?
    
    \item[(ii)] Can we also complete the task in (i), even given corrupted traffic data?
\end{enumerate}

To resolve (i), we extend the deep autoencoder \cite{hinton2006reducing}, a classical unsupervised machine learning approach for dimension reduction, i.e., converting high-dimensional data into latent variables, typically with lower dimensions, by minimizing the error between encoder inputs and decoder outputs; see Figure~\ref{fig_extension}a). Generally, the original low-dimensional representations do not have significant physical meanings; they cannot be recognized as parameters of traffic flow models. To address this problem, we also feed encoder outputs into traffic flow models and inform our decoder of extra discrepancies between the simulation result and its own output; see Figure~\ref{fig_extension}b). Clearly, minimizing the new error encourages the decoder to learn physical laws of traffic flows, and the total error decreases only when the encoder yields appropriate parameters of traffic flow models. Besides, we introduce the concept of conditional generation. That is, the decoder relies not only on latent variables (parameters), but also on boundary conditions, such as upstream traffic volumes. This is a straightforward but indispensable extension since traffic observations, mimicked by the decoder, are determined by these conditions.



Note this paper considers the classical first-order cell transmission model (CTM). Assuredly, high-order models like Payne-Whitham (PW) model \cite{payne1971models,whitham1974linear}, along with its discretized version METANET \cite{messner1990metanet}, and Aw-Rascle-Zhang (ARZ) model \cite{aw2000resurrection,zhang2002non} can reproduce traffic phenomena with higher accuracy, but they have more parameters that could complicate the training process. To the best of our  knowledge, this is the first attempt to estimate traffic flow parameters using unsupervised training. It is thus worthwhile starting with a simple model. We leave identification of high-order models as a future research task. Specially, it will be interesting to investigate whether we can accelerate it via calibrating first-order models since first- and high-order models have common traffic flow parameters.

To answer (ii), we integrate denoising autoencoder \cite{vincent2008extracting}, a simple but robust variant of the original autoencoder, into the calibration approach. That is, we use unspoiled sensor readings to generate new data with partially missing values, which mimics the pattern of real unreliable data, and then apply this synthesized data to training; see Figure~\ref{fig_extension}c). It allows to deploy the parameter identification on real data with missing values after training. Ideally, if there are too many missing values, any method will learn nothing and end up with poor calibration. Thus we also provide a sensitivity analysis by controlling the missing rates to empirically reveal how our approach degenerates.

\subsection{Related work}
Most of the previous work on calibrating macroscopic traffic flow models employed optimization methods applied over specific days. One approach is to estimate the fundamental diagram (FD), especially for first-order traffic flow models \cite{munoz2004methodological,dervisoglu2009automatic,zhong2016automatic}. It requires little computation, but may suffer from accuracy losses. First, fitting the left part of FD, namely free-flow regime, is usually easy, but the same task can be hard for the right part since traffic data collected during congestion periods are sparse and scattered. Besides, individual calibration of FDs along a freeway corridor fails to capture flow interactions. More studies formulated the calibration problem as mathematical programming that considered flow dynamics specified by first- or high-order models \cite{lee2008calibration,spiliopoulou2014macroscopic,mudigonda2015robust,mohammadian2021performance}. In this case, one needs to solve a non-convex optimization problem with locally optimal solutions, and thus heuristic optimization/search algorithms, such as simultaneous perturbation stochastic approximation, simulated annealing, etc, can be used; please see \cite{wang2022macroscopic} for a comprehensive review.

Recently, some researchers discussed the idea of learning-based calibration in the context of traffic state estimation (TSE). That is, they integrated traffic flow models into machine learning methods to enhance TSE \cite{huang2020physics,yuan2021traffic,yuan2021macroscopic,shi2021physics,shi2021physics2}, to incorporate simultaneous parameter identification and state estimation techniques. It should be pointed out that these studies are different from what is proposed in this paper. First, they put emphasis on TSE that used partial observations to infer full states. Thus, they divided training and testing data by sensor locations to evaluate transferability over space. By contrast, our method takes in full observations and returns traffic parameters. We desire transferability over time periods and thereby separate training and testing data by time. Second, although the current studies can update traffic parameters, they still require relatively good initial values that are normally obtained from classical calibration approaches \cite{yuan2020highway} because poor initial traffic parameters do not guarantee  convergence \cite{shi2021physics2}. This kind of good initial parameter estimates, however, is not necessarily required by our approach.

\begin{figure}[htbp]
    \centering
    \includegraphics[width=0.8\linewidth]{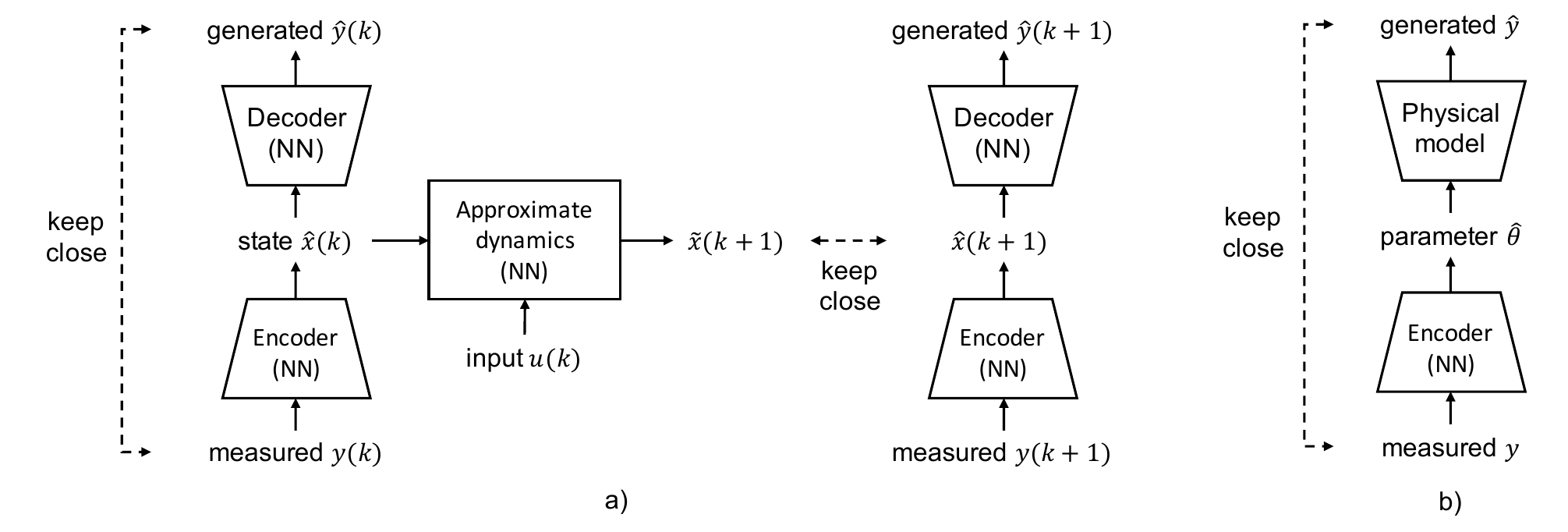}
    \caption{Current frameworks of autoencoder-based system identification: a) identifying approximate dynamics, b) identifying parameters.}
    \label{fig_framework}
\end{figure}

Our proposed method is inspired by deep autoencoder-based system identification which has emerged in recent years. Depending on identification goals, the latent variables, obtained from the autoencoder, can be recognized as either states or parameters. If one wants to fit a dynamical model approximating to the physical one, the encoder gives the states \cite{masti2021learning,beintema2021nonlinear,gedon2021deep}; see the full framework in Figure~\ref{fig_framework}a). If the target is exact parameters of physical models, the encoder can yield calibration results as well; see Figure~\ref{fig_framework}b). In that case the decoder is a physical model rather than a NN. Up to now, this framework has only been  applied to identification of linear time-invariant (LTI) systems \cite{nagel2021autoencoder}. Though it has the same objective as our method, it cannot be directly applied in our problem setting for the following reasons. First, LTI systems have closed-form solutions, and it is convenient to compute gradients with respect to parameters, which implies easy training. However, when  physical models, like traffic flow models, are too complicated to have analytical solutions, gradient calculation becomes very hard by making the training inefficient. Second, it is assumed that step signals are standard inputs for all LTI systems and they are thus ignored in the identification framework. In practice, however, we cannot always manipulate model inputs, i.e., boundary conditions, and should include them in the learning-based calibration.
\section{Problem Statement}
In this section, we generally state the calibration problem  and then explicitly present the considered freeway model. It consists of a dynamics model, the CTM incorporating capacity drop, and an observation model.

\subsection{Learning-based calibration problem}
\begin{figure}[htbp]
    \centering
    \includegraphics[width=0.65\linewidth]{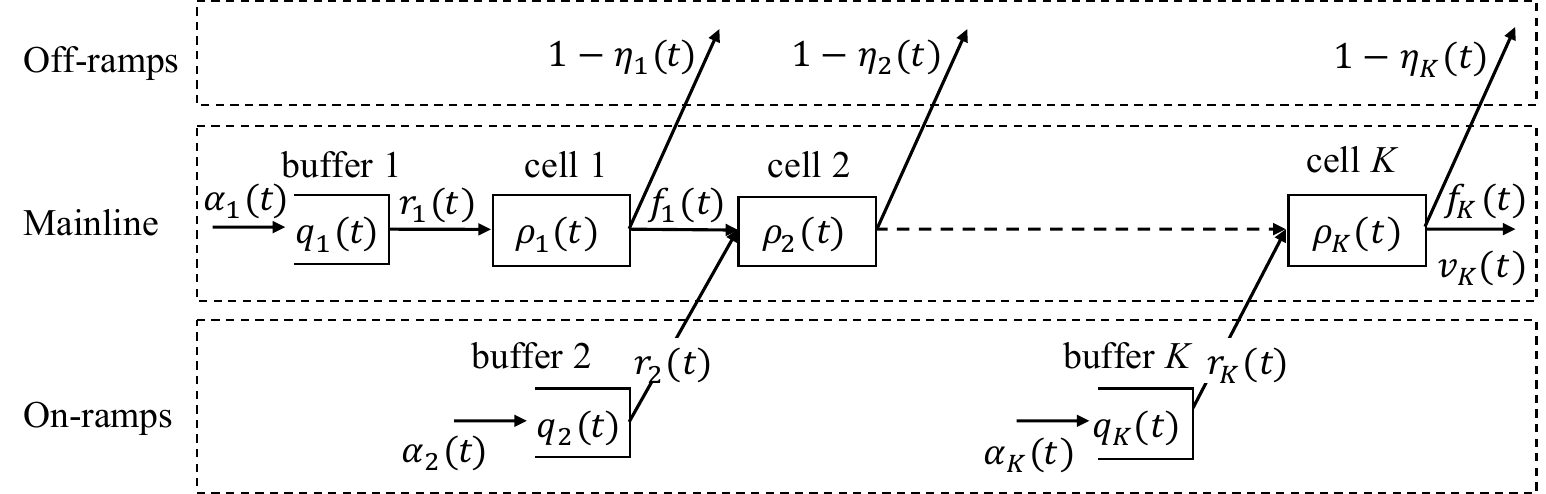}
    \caption{A freeway corridor with states $(\rho(t), q(t))$ and model inputs $u(t):=(\alpha(t), \eta(t), v_K(t))$, where $\rho(t):=[\rho_1(t),\rho_2(t),\cdots,\rho_K(t)]^{\mathrm{T}}$,  $q(t):=[q_1(t),q_2(t),\cdots,q_K(t)]^{\mathrm{T}}$,  $\alpha(t):=[\alpha_1(t),\alpha_2(t),\cdots,\alpha_K(t)]^{\mathrm{T}}$ and $\eta(t):=[\eta_1(t),\eta_2(t),\cdots,\eta_K(t)]^{\mathrm{T}}$.}
    \label{fig_freeway}
\end{figure}

We consider a freeway corridor with $K$ mainline \emph{cells}, $K$ on-ramp \emph{buffers}, and $K$ off-ramps, as shown in Figure~\ref{fig_freeway}. The $k$th buffer has a state of queue length, denoted by  $q_k(t)$, and the $k$th cell is characterized by traffic density, denoted by $\rho_k(t)$. Note that the first buffer is not an actual on-ramp; instead, it represents the upstream freeway section and stores the upstream mainline traffic. The $k$th buffer is subject to a time-varying \emph{demand} $\alpha_k(t)\in\mathbb{R}_{\geq0}$. In addition, we apply \emph{mainline ratio} $\eta_k(t)\in[0, 1]$ to model off-ramp flows. This ratio denotes the fraction of traffic from cell $k$ entering cell $k+1$; the remaining traffic flow leaves the freeway at the $k$th off-ramp. 
We also assume that the last cell $K$ discharges outflows at a speed of $v_K(t)$, which can be measured as the downstream boundary condition.
Finally, we denote by $f_k(t)$ the flows from cell $k$ to the downstream cell and by $r_k(t)$ the flow from buffer $k$ to cell $k$.

Now we suppose that the following freeway model:
\begin{subequations}
    \begin{align}
        \rho(t+1) =& F(\rho(t), q(t), u(t);\theta), \label{eq_rho}\\
        q(t+1) =& G(\rho(t), q(t), u(t);\theta), \label{eq_r} \\
        y(t) =& H(\rho(t),q(t)), \label{eq_idealObs}
    \end{align}
\end{subequations}
where $F$ and $G$ in \eqref{eq_rho}-\eqref{eq_r} denote the dynamics models parameterized by $\theta$, and $H$ in \eqref{eq_idealObs} is the observation model with the ideal senor output $y(t)$, i.e., without missing values. We let $y^{\mathrm{obs}}(t)$ represent real observations of $y(t)$, which can contain missing values. 

Clearly, the traffic flow model assumes constant parameters. Though we admit  variation in traffic parameters, it is still acceptable to assume stationary values over a certain period, i.e., morning or evening peaks \cite{wang2022macroscopic}. This is the basis of our method. Our calibration does not address abrupt parameter changes due to incidents and other unexpected disruptions. This kind of real-time parameter identification that may be solved by online machine learning algorithms is a future research task. For convenience of notation, we let $\theta_p$, $\mathbf{M}_p:= (y^{\mathrm{obs}}(0), y^{\mathrm{obs}}(1), \cdots, y^{\mathrm{obs}}(T))_p$ and $\mathbf{B}_p:=(\rho(0), u(0), u(1), \cdots, u(T))_p$ denote the period-specific parameter, real measurements and boundary conditions, respectively.

The  calibration problem is formulated as follows. Suppose $\mathcal{P}^{\mathrm{train}}$ is a set of periods, each of them with $T+1$ time steps. Given observations $\{\mathbf{M}_p\}_{p\in\mathcal{P}^{\mathrm{train}}}$ and boundary conditions $\{\mathbf{B}_p\}_{p\in\mathcal{P}^{\mathrm{train}}}$, we aim at training a machine learning model so that it can calibrate the model parameters not only on the training data set but also on the testing ones $\mathcal{P}^{\mathrm{test}}$, $\{\mathbf{M}_p\}_{p\in\mathcal{P}^{\mathrm{test}}}$ and $\{\mathbf{B}_p\}_{p\in\mathcal{P}^{\mathrm{test}}}$. Since we do not have true values of $\{\theta_p\}_{p\in\mathcal{P}^{\mathrm{train}}}$ and $\{\theta_p\}_{p\in\mathcal{P}^{\mathrm{test}}}$, we evaluate the calibrated results $\{\hat{\theta}_p\}_{p\in\mathcal{P}^{\mathrm{train}}}$ and $\{\hat{\theta}_p\}_{p\in\mathcal{P}^{\mathrm{test}}}$ by comparing real measurements with those from the model \eqref{eq_idealObs}.

\subsection{Freeway model}
In the following, we first use the CTM to specify the dynamics functions $F$ and $G$, and the parameter $\theta$ in \eqref{eq_rho}-\eqref{eq_r}.  The CTM is favored due to its simplicity and wide use, and it is not a necessary requirement. Then we consider induction loops for the observation function $H$ and the sensor output $y(t)$ in \eqref{eq_idealObs}. Although Lagrangian sensors, such as floating cars, with higher accuracy have been previously studied \cite{work2009lagrangian, yuan2012real}, their observations could be sparse given low market penetration rates. Besides, there is limited public access to them. By contrast, we have rich sources of induction loop data, which supports training and validation of learning-based models.

\subsubsection{Dynamics model}
The flow from buffer $k$ to cell $k$, $r_k$, is specified by
\begin{equation}
    r_k(t) = \min\{\alpha_k(t)+q_k(t)/\delta_t,U_k, w_k(\rho^{\max}_k-\rho_k(t))\},
\end{equation}
where $\delta_t$ denotes time step size, $U_k$ denotes capacity of buffer $k$, $\rho_{k}^{\max}$ denotes jam density of cell $k$, and $w_k$ denotes congestion wave speed of cell $k$. Then the flows between cells, $f_k$ for $k=1,2,\cdots,K$, are given by
\begin{subequations}
\begin{align}
    &f_k(t) =\eta_k(t)\min\big\{v_k\rho_k(t), Q_k(t),w_{k+1}(\rho^{\max}_{k+1}-\rho_{k+1}(t))-r_{k+1}(t)\big\}, ~1\leq k \leq K-1 \label{eq_fk} \\
    &f_K(t) = v_K(t)\rho_K(t) \label{eq_fK}
\end{align}
\end{subequations}
where $\delta_t$ denotes time step size, $v_k$ denotes free-flow speed of cell $k$, $Q_k(t)$ denotes capacity of cell $k$, $\rho_{k}^{\max}$ denotes jam density of cell $k$, and $w_k$ denotes congestion wave speed of cell $k$. The flow functions \eqref{eq_fk}-\eqref{eq_fK} indicate higher merging priority of on-ramp flows and the first-in-first-out rule for off-ramp flows \cite{ferrara2018freeway}. Besides, note that we consider time-varying capacity $Q_k(t)$ which allows to model capacity drop as follows:
\begin{equation}
    Q_k(t) = \begin{cases}
    Q^{\mathrm{nominal}}_k, & \rho_k(t)\leq \rho_k^{\mathrm{critical}}, \\
    Q^{\mathrm{drop}}_k, &  \rho_k(t) > \rho_k^{\mathrm{critical}},
    \end{cases}
\end{equation}
where $\rho_k^{\mathrm{critical}}:=Q_k^{\mathrm{nominal}}/v_k$ denotes critical density of cell $k$; see more discussions and implementations of capacity drop in \cite{kontorinaki2017first}.

Then, by the conservation law of flows, the traffic dynamics is given by
\begin{subequations}
\begin{align}
    q_k(t+1) =& q_k(t)  + \delta_t(\alpha_k(t) - r_k(t)),~1\leq k \leq K,~t=0, 1, \cdots, \label{eq_queuedynamics} \\
    \rho_1(t+1) =& \rho_1(t) + \frac{\delta_t}{\ell_1}(r_1(t)-\frac{f_1(t)}{1-\eta_1(t)}),~t=0, 1, \cdots, \\
    \rho_k(t+1) =& \rho_k(t) + \frac{\delta_t}{\ell_k}(r_k(t)+f_{k-1}(t)-\frac{f_k(t)}{1-\eta_k(t)}),~2\leq k \leq K,~t=0, 1, \cdots,
\end{align}
\end{subequations}
where $\ell_k$ denotes the length of cell $k$. 

Note that the model above assumes infinite-sized buffers. It helps to store boundary inflows given insufficient cell space that is probably caused by the selection of inappropriate parameters during the training process. We also introduce $U_k$, $k=1,2,\cdots,K$, to prevent unrealistically large inflows when there are long queues at buffers. All of these parameters can be specified in advance. Then the parameters to be calibrated are presented below: 
$$\theta = (\{v_k\}_{k=1}^{K-1}, \{Q_k^{\mathrm{nominal}}\}_{k=1}^{K-1}, \{Q_k^{\mathrm{drop}}\}_{k=1}^{K-1} \{\rho_k^{\max}\}_{k=1}^{K}, \{w_k\}_{k=1}^{K}).$$

\subsubsection{Observation model}
In practice, induction loops update measurements of flow rates and speed at a certain frequency $\Delta_t$ that is larger than the time step size $\delta_t$ of the traffic model. We suppose $\Delta_t=m\delta_t$ with a multiple $m\in\mathbb{Z}_{>0}$. Then the sensor outputs are given by
\begin{subequations}
\begin{align}
    \bar{r}_k(t) =& \sum_{i=t-m+1}^{t} r_k(i)/m,~1\leq k \leq K,~t=m, 2m, ..., \label{eq_obs1} \\
    \bar{f}_k(t) =& \sum_{i=t-m+1}^{t} f_k(i)/m,~1\leq k \leq K,~t=m, 2m, ..., 
    \label{eq_obs2}
    \\
    \bar{v}_k(t) =& \sum_{i=t-m+1}^{t} f_k(i)v_k(i)/\sum_{i=t-m+1}^{t} f_k(i),~1\leq k \leq K,~t=m, 2m, ...,
    \label{eq_obs3}
\end{align}
\end{subequations}
where $v_k(t) := f_k(t)/\rho_k(t)$ denotes traffic speed of cell $k$ at time $t$. Clearly, \eqref{eq_obs1}-\eqref{eq_obs3}  yield
$$y(t)=[\bar{r}_1(t), \bar{r}_2(t),\cdots,\bar{r}_K(t),\bar{f}_1(t), \bar{f}_2(t),\cdots,\bar{f}_K(t), \bar{v}_1(t), \bar{v}_2(t),\cdots,\bar{v}_K(t)]^{\mathrm{T}}, t=m,2m,\cdots.$$

\section{Proposed Method}
In this section, we first introduce the autoencoder-based parameter identification that assumes complete traffic measurements and then present its extension for corrupted data.

\subsection{Autoencoder-based parameter identification}
At first, the encoder outputs the estimated parameter $\hat{\theta}_p$ given the measurement $\mathbf{M}_p$ and the boundary condition $\mathbf{B}_p$. Then the parameter, along with the boundary condition, is passed into the encoder and the CTM to obtain $\hat{\mathbf{M}}_p$ and $\tilde{\mathbf{M}}_p$, respectively. The computations are presented below:
\begin{subequations}
\begin{align}
    \hat{\theta}_p =& E(\mathbf{M}_p, \mathbf{B}_p; w_E), \label{eq_encoder} \\
    \hat{\mathbf{M}}_p =& D(\hat{\theta}_p, \mathbf{B}_p; w_D), \\
    \tilde{\mathbf{M}}_p =& C(\hat{\theta}_p,\mathbf{B}_p), 
\end{align}
\end{subequations}
where $E$, $D$, $C$ denote the encoder, the decoder and the CTM. Note that both the encoder and the decoder are neural networks, parameterized by weights $w_E$ and $w_D$ respectively. The estimated parameters $\hat{\theta}_p$ should fall within feasible ranges; otherwise it cannot be passed into the CTM. Thus we exploit the sigmoid function for rescaling in the last layer of the encoder as follows:
\begin{equation}
    \hat{\theta}_p = (\theta^{\max} - \theta^{\min})\odot\mathrm{sigmoid}(\hat{\theta}^{\mathrm{unscaled}}_p)+ \theta^{\min}
\end{equation}
where $\odot$ denotes element-wise product, $\hat{\theta}_p^{\mathrm{unscaled}}$ is the output from the previous layer, and $\theta^{\mathrm{min}}$ (resp. $\theta^{\mathrm{max}}$) is a lower bound (resp. an upper bound) of traffic parameters.

As for training, we consider minimizing the following loss function:
\begin{equation}
    \min_{w_E,w_D} L = \min_{w_E,w_D} \sum_{p\in\mathcal{P}^{\mathrm{train}}}||\hat{\mathbf{M}}_p-\tilde{\mathbf{M}}_p||_2^2 + \gamma\sum_{p\in\mathcal{P}^{\mathrm{train}}} ||\mathbf{M}_p-\hat{\mathbf{M}}_p||_2^2 \label{eq_loss}
\end{equation}
where $\gamma$ is a loss weight. Clearly, the loss function includes two parts. Minimizing the first part ensures that the decoder learns the physical laws of the CTM, and then minimizing the second part induces the encoder to yield appropriate traffic flow parameters.

\subsection{Denoising autoencoders-based parameter identification}
Considering the observation $\mathbf{M}_p$ could have missing values, we define a binary matrix $\mathbf{I}_p$, with the same dimension as $\mathbf{M}_p$, to indicate whether the corresponding data is missing. We first sample from the set of complete observations and then from $\{\mathbf{I}_p\}_{p\in\mathcal{P}^{\mathrm{train}}}$ to assemble a new observation $\mathbf{M}_p^*$. Although it may have missing values, their corresponding ground truths are known. We feed the artificial observation $\mathbf{M}_p^*$ to the encoder. That is, instead of \eqref{eq_encoder}, we use the following
\begin{equation}
    \hat{\theta}_p = E(\mathbf{M}_p^*, \mathbf{B}_p; w_E)
\end{equation}
but we still apply \eqref{eq_loss} to training.
\section{Experimental Design}

\subsection{Data preparation}
We consider the freeway segment, up to 6.2 kilometers, shown in Figure~\ref{fig_motivating1}. It has 6 on-ramps and 4 off-ramps. We divide it into 9 cells based on locations of on-ramps, off-ramps and sensors. We also collected 5-min traffic flow and speed data from 2017 to 2019 via the PeMS \cite{chen2002freeway}. The preliminary data analysis showed recurrent morning and evening peaks. Thus we separately calibrate for these two periods for each day. We select the data of 2017 and 2018 as the training dataset and the data of 2019 as the testing dataset.

\subsection{Preliminary results}
\begin{figure}[h]
    \centering
    \includegraphics[width=0.5\linewidth]{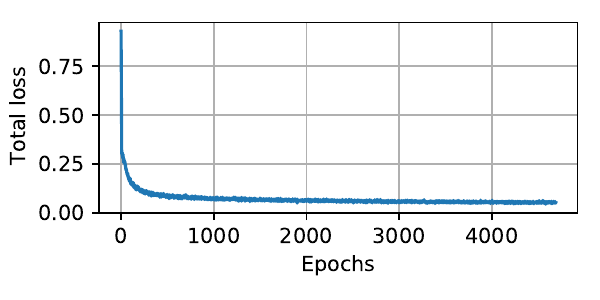}
    \caption{Training loss of learning-based calibration given intact sensor data.}
    \label{fig_trainingloss}
\end{figure}

We have trained our machine learning model for the calibration given complete sensor measurements. We set the structure of neural networks, as the encoder and the decoder, based on the LeNet \cite{lecun1998gradient} with customized input and output layer sizes. We also selected the loss weight $\gamma=0.8$. Figure~\ref{fig_trainingloss} presents the training loss.

\begin{figure}[htbp]
    \centering
    \includegraphics[width=0.8\linewidth]{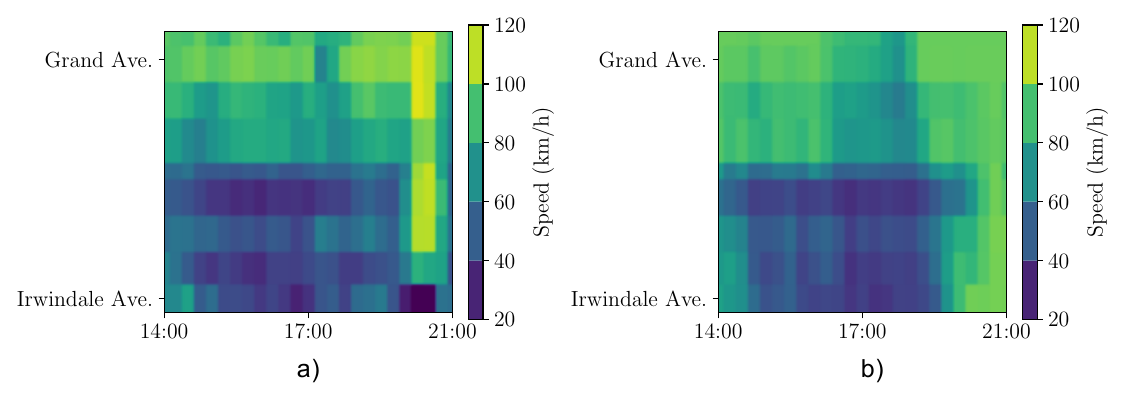}
    \caption{Speed evaluation on March 26, 2019: a) ground truth, b) simulation results.}
    \label{fig_evalspeed}
\end{figure}

\begin{figure}[htbp]
    \centering
    \includegraphics[width=0.8\linewidth]{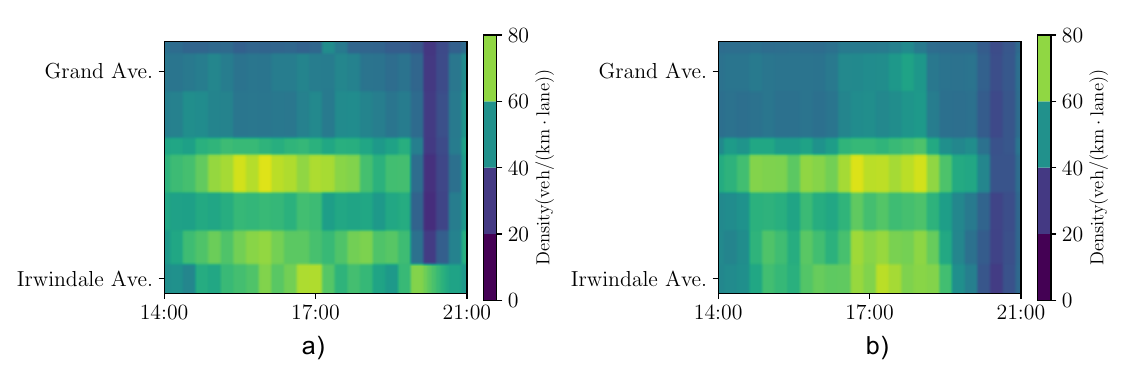}
    \caption{Traffic density evaluation on March 26, 2019: a) ground truth, b) simulation results.}
    \label{fig_evalden}
\end{figure}

We tested the trained model on March 26, 2019. For that, we first used the encoder to yield the parameters required by the CTM and then applied the traffic flow model to numerical simulation. Figure~\ref{fig_evalspeed} and \ref{fig_evalden} presented the comparison in terms traffic speed and density, respectively. These results showed that our machine learning model can yield reasonable calibration so that the traffic flow model can reproduce the congestion pattern.

\section{Summary and Future Work}
In this paper, we propose a physics-informed, learning-based calibration approach, inspired by autoencoders comprising of one encoder and one decoder, that is expected to achieve performances comparable to those of optimization-based methods while not requiring day-to-day optimization after training. We consider calibrating the CTM, a widely-used traffic flow model. In our approach, the encoder takes as input traffic measurements and boundary conditions, and yields parameters required by CTM; the decoder recovers the measurements from the encoder output and the boundary conditions. Specially, we feed the decoder input to CTM and inform the autoencoder of a novel error between the decoder output and the simulation results besides the conventional error between the traffic measurements and the decoder output. This encourages the encoder to produce reasonable parameters so that the new error is minimized. We also introduce the denoising autoencoder into our calibration method so that it can handles with corrupted data. It is expected to demonstrate the effectiveness of the proposed method through a case study of I-210 E in California.

We are currently continuing to train our machine learning model to test the calibration with corrupted data. In the full version of the paper, we will also compare our method with several benchmarks. We will first consider the case of complete sensor measurements. We will then apply case-by-case optimization on the testing data, one fitting the fundamental diagrams \cite{zhong2016automatic} and the other one  solving non-convex optimization \cite{spiliopoulou2014macroscopic}. Then, for the case of corrupted data, we will test these two methods for different scenarios, one only with raw data and the other one with imputed data.  

\bibliographystyle{IEEEtran}
\bibliography{Bibliography}

\end{document}